\documentclass{article} 
\usepackage{iclr2021_conference,times}

\usepackage{amsmath,amsfonts,bm}

\def\eqref#1{equation~\ref{#1}}

\def\1{\bm{1}}

\DeclareMathAlphabet{\mathsfit}{\encodingdefault}{\sfdefault}{m}{sl}
\SetMathAlphabet{\mathsfit}{bold}{\encodingdefault}{\sfdefault}{bx}{n}

\usepackage{graphicx}
\usepackage{booktabs}
\usepackage{hyperref}
\usepackage{url}
\usepackage{textcomp}
\usepackage{array,multirow,graphicx}
\usepackage{float}
\newcommand{\eat}[1]{}
\title{Graph Autoencoders with Deconvolutional Networks}

\eat{
\author{Antiquus S.~Hippocampus, Natalia Cerebro \& Amelie P. Amygdale \thanks{ Use footnote for providing further information
about author (webpage, alternative address)---\emph{not} for acknowledging
funding agencies.  Funding acknowledgements go at the end of the paper.} \\
Department of Computer Science\\
Cranberry-Lemon University\\
Pittsburgh, PA 15213, USA \\
\texttt{\{hippo,brain,jen\}@cs.cranberry-lemon.edu} \\
\And
Ji Q. Ren \& Yevgeny LeNet \\
Department of Computational Neuroscience \\
University of the Witwatersrand \\
Joburg, South Africa \\
\texttt{\{robot,net\}@wits.ac.za} \\
\AND
Coauthor \\
Affiliation \\
Address \\
\texttt{email}
}
}
\author{Jia Li$^1$, Tomas Yu$^1$, Da-Cheng Juan$^2$, Arjun Gopalan$^2$, 
\textbf{Hong Cheng}$^1$\textbf{,}  \textbf{Andrew Tomkins}$^2$\\
$^1$ The Chinese University of Hong Kong\\
$^2$ Google Research\\
\texttt{\footnotesize \{lijia,hcheng\}@se.cuhk.edu.hk}\\ 
\texttt{\footnotesize \{dacheng,arjung,tomkins\}@google.com }
}

\iclrfinalcopy 
\begin{document}

\maketitle

\begin{abstract}
Recent studies have indicated that Graph Convolutional Networks (GCNs) act as a \emph{low pass} filter in spectral domain and encode smoothed node representations.  In this paper, we consider their opposite, namely Graph Deconvolutional Networks (GDNs) that reconstruct graph signals from smoothed node representations. We motivate the design of Graph Deconvolutional Networks via a combination of inverse filters in spectral domain and de-noising layers in wavelet domain, as the inverse operation results in a \emph{high pass} filter and may amplify the noise.  Based on the proposed GDN, we further propose a graph autoencoder framework that first encodes smoothed graph representations with GCN and then decodes accurate graph signals with GDN. 
We demonstrate the effectiveness of the proposed method on several tasks including unsupervised graph-level representation , social recommendation and graph generation. 
\end{abstract}

\section{Introduction}
\label{headings}

Autoencoders have demonstrated excellent performance on tasks such as unsupervised representation learning \citep{bengio2009learning} and de-noising \citep{vincent2010stacked}. Recently, several studies \citep{zeiler2014visualizing,long2015fully} have demonstrated that the performance of autoencoders can be further improved by encoding with Convolutional Networks and decoding with Deconvolutional Networks \citep{zeiler2010deconvolutional}.
Notably, \citet{noh2015learning} present a novel symmetric architecture that provides a bottom-up mapping from input signals to latent hierarchical feature space with \{convolution, pooling\} operations and then maps the latent representation back to the input space with \{deconvolution, unpooling\} operations.
While this architecture has been successful when processing features with structures existed in the Euclidean space (e.g., images), recently there has been a surging interest in applying such a framework on non-Euclidean data like graphs.
However, extending this autoencoder framework to graph-structured data requires Graph Deconvolutional operations, which remains open-ended and hasn't been well-studied as opposed to the large body of works that have already been proposed for Graph Convolutional Networks \citep{defferrard2016convolutional,kipf2017semi}. 
In this paper, we study the characteristics of Graph Deconvolutional Networks (GDNs), and observe de-noising to be the key for effective deconvolutional operations. Therefore, we propose a wavelet-based module \citep{hammond2011wavelets} that serves as a de-noising mechanism after the signals reconstructed in the spectral domain \citep{shuman2013emerging} for deconvolutional networks.

Most GCNs proposed by prior arts, e.g., Cheby-GCN \citep{defferrard2016convolutional} and GCN \citep{kipf2017semi}, exploit spectral graph convolutions \citep{shuman2013emerging} and Chebyshev polynomials \citep{hammond2011wavelets} to retain coarse-grained information and avoid explicit eigen-decomposition of the graph Laplacian. 
Until recently, \citet{pmlr-v97-wu19e} and \citet{donnat2018learning} have noticed that GCN acts as a \emph{low pass} filter in spectral domain and retains smoothed representations.
Inspired by prior arts in the domain of signal deconvolution \citep{kundur1996blind}, we propose to design a GDN by using \emph{high pass} filters as the counterpart of \emph{low pass} filters embodied in GCNs.
Due to the nature of signal deconvolution being ill-posed, several prior arts \citep{donoho1994ideal,figueiredo2003algorithm} rely on transforming these signals into another domain (e.g., spectral domain) where the problem can be better posed and resolved.
Furthermore, \citet{neelamani2004forward} observe inverse filters in spectral domain may amplify the noise, and we observe the same phenomenon for GDNs. Therefore, inspired by their proposed hybrid spectral-wavelet method---inverse signal reconstruction in spectral domain followed by a de-noising step in wavelet domain---we introduce a spectral-wavelet GDN to decode the smoothed representations into the input graph signals. The proposed spectral-wavelet GDN employs spectral graph convolutions with a \emph{high pass} filter to obtain inversed signals and then de-noises the inversed signals in wavelet domain. In addition, we apply Maclaurin series as a fast approximation technique to compute both \emph{high pass} filters and wavelet kernels \citep{donnat2018learning}.

With the proposed spectral-wavelet GDN, we further propose a graph autoencoder (GAE) framework that resembles the symmetric fashion of architectures \citep{noh2015learning}. We then evaluate the effectiveness of the proposed GAE framework with three popular and important tasks: unsupervised graph-level representation \citep{sun2019infograph}, social recommendation \citep{jamali2010matrix} and graph generation. In the first task, the proposed GAE outperforms the state-of-the-arts on graph classification in an unsupervised fashion, along with a significant improvement on running time. In the second task, the performance of our proposed GAE is on par with the state-of-the-arts on the recommendation accuracy; at the meantime, the proposed GAE demonstrates strong robustness against rating noises and achieves the best recommendation diversification \citep{ziegler2005improving}. In the third task, our proposed GDN can enhance the generation performance of popular variational autoencoder frameworks including VGAE \citep{kipf2016variational} and Graphite \citep{grover2018graphite}.

\section{Related work}\label{sec.rel}
\paragraph{Deconvolutional networks} The area of signal deconvolution \citep{kundur1996blind} has a long history in the signal processing community and is about the process of estimating the true signals given the degraded or smoothed signal characteristics \citep{banham1997digital}. Later deep learning studies \citep{zeiler2010deconvolutional,noh2015learning} consider deconvolutional networks as the opposite operation for Convolutional Neural Networks (CNNs) and have mainly focused on Euclidean structures, e.g., image. Some work \citep{dumoulin2016guide} notices \citet{zeiler2010deconvolutional} is in essence a transposed convolution network as it differs from what is used in the signal processing community. For deconvolutional networks in non-Euclidean structures like graphs, the study is still sparse.  \citet{feizi2013network} propose the network deconvolution as inferring the true network given partially observed structure. It relies on explicit eigen-decomposition and cannot be used as the counterpart for GCN. \citet{yang2018enhancing} formulate the deconvolution as a pre-processing step on the observed signals, in order to improve classification accuracy. \citet{zhang2020graph} consider recovering graph signals from the latent representation. However, it just adopts the filter design used in GCN and sheds little insight into the internal operation of GDN.
\paragraph{Graph autoencoders}
Since the introduction of Graph Neural Networks (GNNs) \citep{kipf2017semi,defferrard2016convolutional} and autoencoders (AEs), many studies \citep{kipf2016variational,grover2018graphite} have used GNNs and AEs to encode to and decode from latent representations. Recently \emph{graph pooling} has emerged as a research topic that also contributes to the development of graph autoencoders. Common practices include DIFFPOOL \citep{ying2018hierarchical}, SAGPool \citep{lee2019self}, MinCut-Pool \citep{bianchi2020mincutpool}.  Although some encouraging progress has been achieved, there is still no work about graph deconvolution that can up-sample latent feature maps to restore their original resolutions \citep{gao2019graph}. In this regard, current graph autoencoders bypass the difficulty via (1) non-parameterized decoders \citep{kipf2016variational,deng2019graphzoom, li2020heatts}, (2) GCN decoders \citep{grover2018graphite,gao2019graph}, and (3) multilayer perceptron (MLP) decoders \citep{simonovsky2018graphvae}.

\begin{figure}
\includegraphics [width=0.95\textwidth]{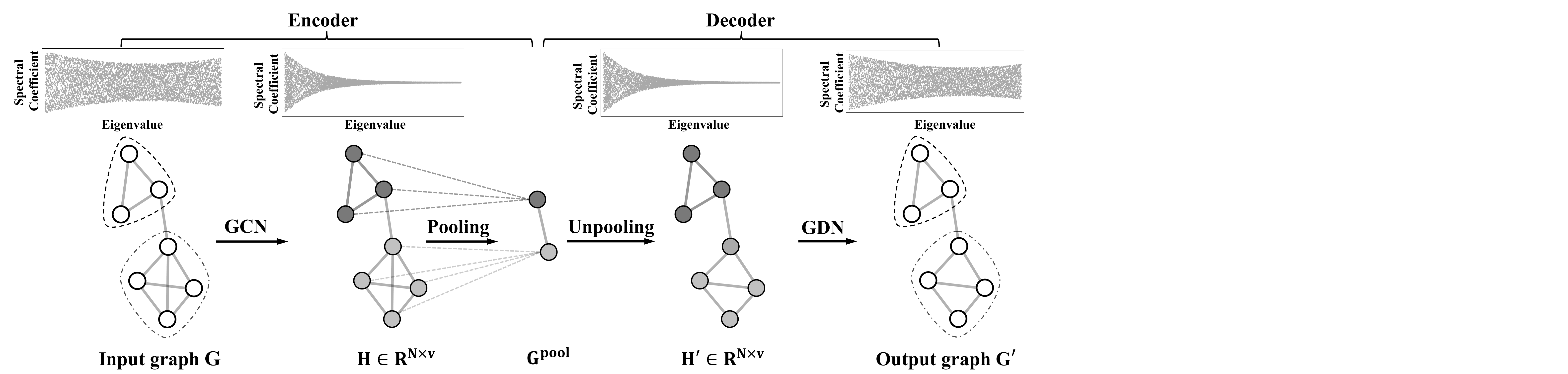}
\caption{Schematic diagram of the proposed graph autoencoder framework. We plot graphs in spatial domain together with their signal coefficients in spectral domain. In the encoding stage, we run a GCN model to derive smoothed node representations and obtain coarse-grained graph via a pool layer. In the decoding stage, we upscale the coarsened graph back and reconstruct the graph signal via a GDN model. }
\label{frame}
\end{figure}

\section{Graph Autoencoder Framework}
\label{headings}

Formally, we are given an undirected, unweighted graph $G = (V,A,X)$. $V$ is the node set and $N = |V|$ denotes the number of nodes. The adjacency matrix $A \in \mathbb{R}^{N\times N}$ represents the graph structure. The feature matrix $X \in \mathbb{R}^{N\times d}$ represents the node attributes. Our goal is to learn an encoder and a decoder to map between the space of graph $G$ and their latent factors $G^{pool} =(V^{pool},A^{pool},Z)$. We show a schematic diagram of our proposed framework in Figure \ref{frame}. 

\subsection{Encoder}\label{en}
Our encoder consists of several layers of Graph Convolutional Networks (GCNs) \citep{kipf2017semi} and a pooling layer, to produce coarser representations of the input graphs.
\paragraph{Convolution}
The convolutional layers are used to derive smoothed node representations, such that nodes that are similar in topological space should be close enough in Euclidean space.
\begin{equation}
 H = \text{GCN}(A,X),
\label{equ.H}
\end{equation}
where $H \in \mathbb{R}^{N\times v}$ denotes smoothed node representations. Specifically, \citet{pmlr-v97-wu19e} show that GCN is a \emph{low pass} filter in spectral domain with $g_c(\lambda_i) = 1 - \lambda_i$, where $\{\lambda_i\}_{i=1}^N$ are the eigenvalues of the normalized Laplacian matrix $L_{sym} = D^{-\frac{1}{2}}LD^{-\frac{1}{2}}$, $L$ and $D$ are the Laplacian and degree matrices of the input graph $A$ respectively. 
\paragraph{Pooling}
We follow \citet{lee2019self} and \citet{jiawww19} by using the self attention mechanism to pool the fine-grained graph into coarse-grained representations,
\begin{equation}
  S = \text{softmax} \big(\text{tanh}(HW_{1})W_{2}\big),
\label{equ.gene}
\end{equation}

where $W_{1} \in \mathbb{R}^{v \times d}$ and $W_{2} \in \mathbb{R}^{d \times K}$ are two weight matrices, $W_{1}$ is used for feature transformation and $W_{2}$ is used to infer the membership of each node with respect to each cluster $V_k$. 

Similar to \citet{ying2018hierarchical}, we compute the coarsed graph structure $A^{pool} \in \mathbb{R}^{K \times K}$ and feature representation $Z \in \mathbb{R}^{K \times v}$ as follows:
\begin{equation}
Z = S^{\top}H;\ \ \  \ \ \ A^{pool} = S^{\top}AS.
\end{equation}
Note here $(Z,A^{pool})$ is size invariant and permutation invariant, as pointed out by \citet{jiawww19}.

\subsection{Decoder}\label{de}
Our decoder consists of an unpooling layer and several layers of Graph Deconvolutional Networks (GDNs), to produce fine-grained graphs from the encoded $G^{pool}$.
\paragraph{Unpooling}
We follow \citet{bianchi2020mincutpool} to upscale the coarsened graph back to its original size,
\begin{equation}
H' = SZ;\ \ \  \ \ \ A' = SA^{pool}S^{\top}.
\end{equation}
\paragraph{Deconvolution}\label{decon}
The deconvolutional layers are used to recover the original graph features given smoothed node representations,
\begin{equation}
 X' = \text{GDN}(A',H'),
\label{equ.X}
\end{equation}
where $X' \in \mathbb{R}^{N\times d}$ denotes the recovered graph features. We shall further discuss our design of GDN in Section \ref{gdns}.
\subsection{The loss function}\label{tloss}
The overall reconstruction loss is a weighted sum of structure reconstruction loss and feature reconstruction loss.
\begin{equation}
 \mathcal{L} = \lambda_Af(A,A') + \lambda_Xf(X,X'),
\end{equation}
where $f(\cdot,\cdot)$ denotes a differential distance metric, e.g., $f(\cdot,\cdot) =  \text{MSE}(\cdot,\cdot)$ for continuous input, and $\text{MSE}(\cdot,\cdot)$ represents mean squared error.
\section{Graph Deconvolutional Networks}
\label{gdns}
In this section, we present our design of Graph Deconvolutional Networks (GDNs). A naive deconvolutional nets can be obtained using the inverse operator $g_c^{-1}(\lambda_i) = \frac{1}{1 - \lambda_i}$ in spectral domain. Unfortunately, inverse operation results in a \emph{high pass} filter and may amplify the noise \citep{donoho1994ideal,figueiredo2003algorithm}. In this regard, we propose an efficient, hybrid spectral-wavelet deconvolutional network that performs inverse signal recovery in spectral domain first, and then conducts a de-noising step in wavelet domain to remove the amplified noise \citep{neelamani2004forward}.

\subsection{Inverse of GCN}\label{inv}
In order to recover graph signals from the latent representation computed by GCN encoder, we proposed a naive approach--inverse GCN with the inverse filter as  $g_c^{-1}(\lambda_i) = \frac{1}{1 - \lambda_i}$ in spectral domain. 
The spectral graph convolution on a signal $x \in \mathbb{R}^N$ is defined as:
\begin{equation}
    \begin{array}{ll}
        g_{c}^{-1} * x = U\text{diag}(g_c^{-1}(\lambda_1),\ldots,g_c^{-1}(\lambda_N))U^\top x = Ug_c^{-1}(\Lambda)U^\top x,
    \end{array}
\end{equation}
where $U$ is the eigenvector matrix of the normalized graph Laplacian $L_{sym} = U{\Lambda}U^\top$. 
Then, we apply Maclaurin series approximation on $g_c^{-1}(\Lambda) = \sum_{n=0}^{\infty}\Lambda^n$ and get a fast algorithm as below: 
\begin{equation}
    \begin{array}{ll}
       g_{c}^{-1} * x = U\sum_{n=0}^{\infty}\Lambda^{n}U^\top x
       =\sum_{n=0}^{\infty}L_{sym}^n x.
    \end{array}
\end{equation}
As in GCN \citep{kipf2017semi}, when the first order approximation is used to address overfitting, we derive a spectral filter with $g_c^{-1}(\lambda_i) = 1 + \lambda_i$, which is apparently a \emph{high pass} filter. 

Following GCN \citep{kipf2017semi}, a feature transformation is applied to increase filter strength. Recap the GDN in Section \ref{decon},
the inverse version of GCN can be written as:
\begin{equation}
 M = (I_N+L'_{sym})H'W_3,
\label{equ.m}
\end{equation}
where $L'_{sym}$ is the corresponding normalized graph Laplacian matrix for $A'$, $H'$ is the smoothed representations and $W_3$ is the parameter set to be learned.

Compared with directly using GCN for signal reconstruction in \citet{zhang2020graph}, the proposed inverse GCN demonstrates its efficacy in recovering the high frequency signals of the graph, as shown in Figure 2 (b) and (d). We shall further discuss this point in Section \ref{rgd}.

\subsection{Wavelet de-noising}\label{wav}
The proposed inverse GCN  may over-amplifies the high frequency signals and introduce undesirable noise into the output graph. 
Thus, a de-noising process is necessary to separate the useful information and the noise. 
Wavelet-based methods have a strong impact on the field of de-noising, especially in image restoration when noise is amplified by inverse filters \citep{neelamani2004forward}. Notably, coefficients in wavelet domain could allow the noise to be easily separated from the useful information, while transformation into spaces such as spectral domain does not share the same characteristics. In the literature, many wavelet de-noising methods have been proposed, e.g., SureShrink \citep{donoho1994ideal}, BayesShrink \citep{chang2000adaptive}, and they differ in how they estimate the separating threshold. Our method generalizes these threshold ideas and automatically separates the noise from the useful information with a learning-based approach.

\begin{figure}
\centering
\includegraphics [width=0.4\textwidth]{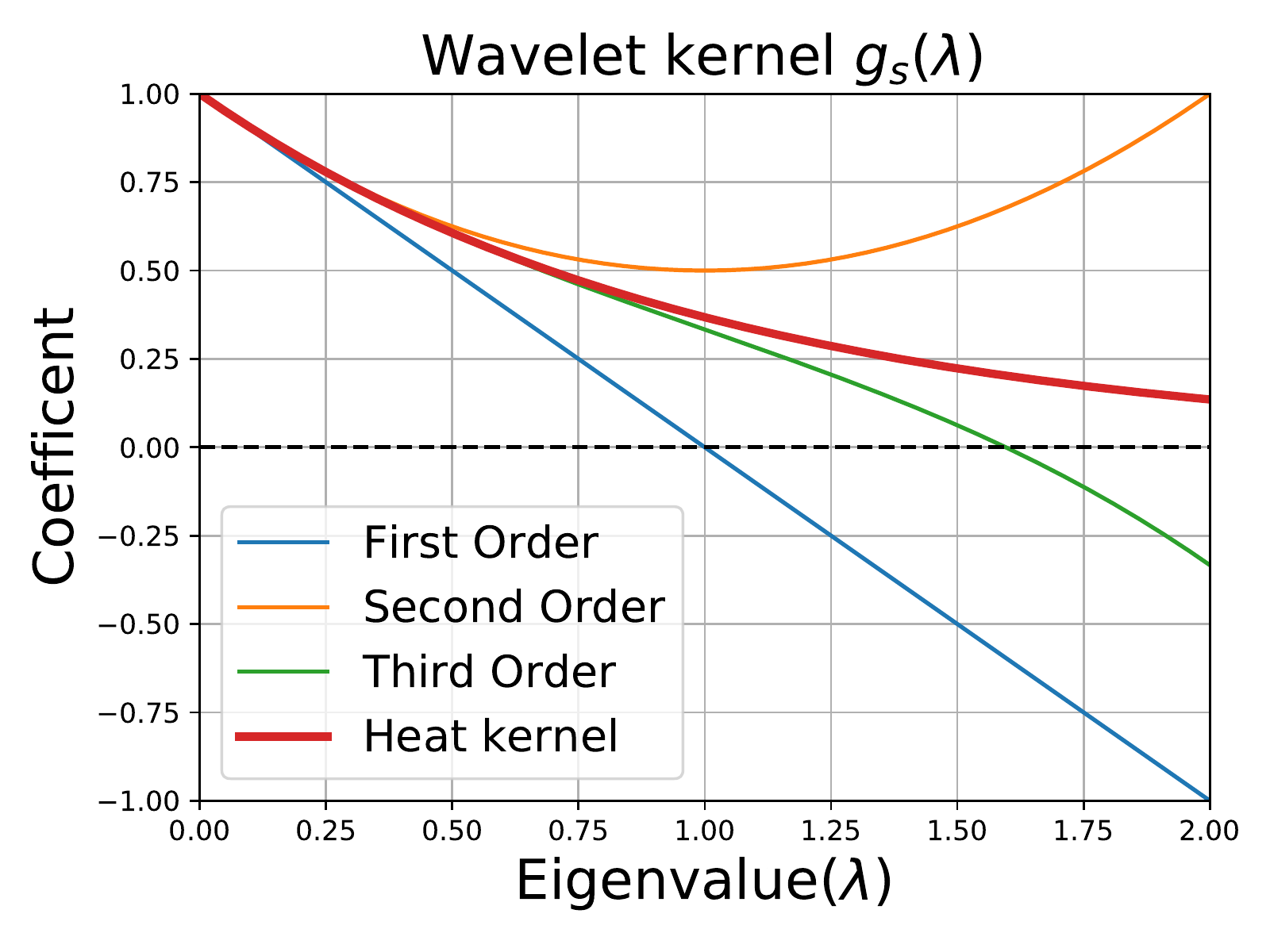}
\includegraphics [width=0.4\textwidth]{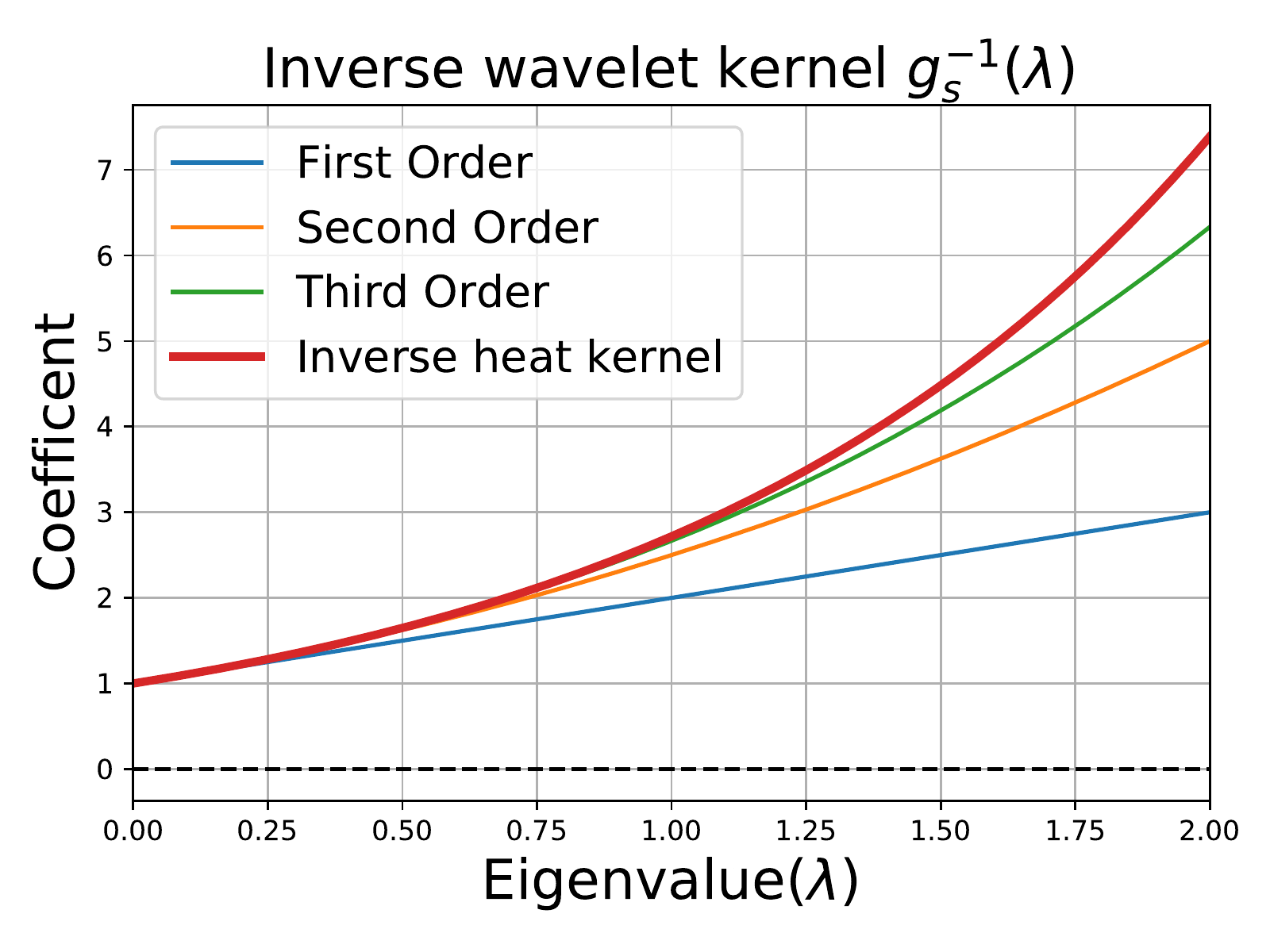}
\caption{Low-order Maclaurin Series well approximate wavelet kernel and inverse wavelet kernel with $s=1$.}
\label{loworder}
\vspace{-0.3cm}
\end{figure}

Consider a set of wavelet bases $\Psi_s = (\Psi_{s1},\ldots,\Psi_{sN})$, where each one $\Psi_{si}$ denotes a signal on graph diffused away from node $i$ and $s$ is a scaling parameter \citep{xu2019graph}, the wavelet bases can be written as $\Psi_s = Ug_s(\Lambda)U^{\top}$ and $g_s(\cdot)$ is a filter kernel. Following previous works GWNN \citep{xu2019graph} and GRAPHWAVE \citep{donnat2018learning}, we use the heat kernel $g_s(\lambda_i) = e^{-s\lambda_i}$, as heat kernel admits easy calculation of inverse wavelet transform with $g_s^{-1}(\lambda_i) = e^{s\lambda_i}$. In addition, we can apply Maclaurin series approximation on heat kernel and neglect explicit eigendecomposition by:

\begin{equation}
    \begin{array}{ll}
       \Psi_s = U\sum_{n}\frac{(-1)^{n}}{n!}s^n\Lambda^{n}U^\top 
       =\sum_{n=0}^{\infty}\frac{(-1)^{n}}{n!}s^nL_{sym}^n,
    \end{array}
\end{equation}

\begin{equation}
    \begin{array}{ll}
       \Psi_s^{-1} = U\sum_{n}\frac{s^n}{n!}\Lambda^{n}U^\top 
       =\sum_{n=0}^{\infty}\frac{s^n}{n!}L_{sym}^n,
    \end{array}
\end{equation}
Usually truncated low order approximation is used in practice with $n\leq 3$ \citep{defferrard2016convolutional}.
\begin{figure}[h]
\begin{center}
\includegraphics [width=0.7\textwidth]{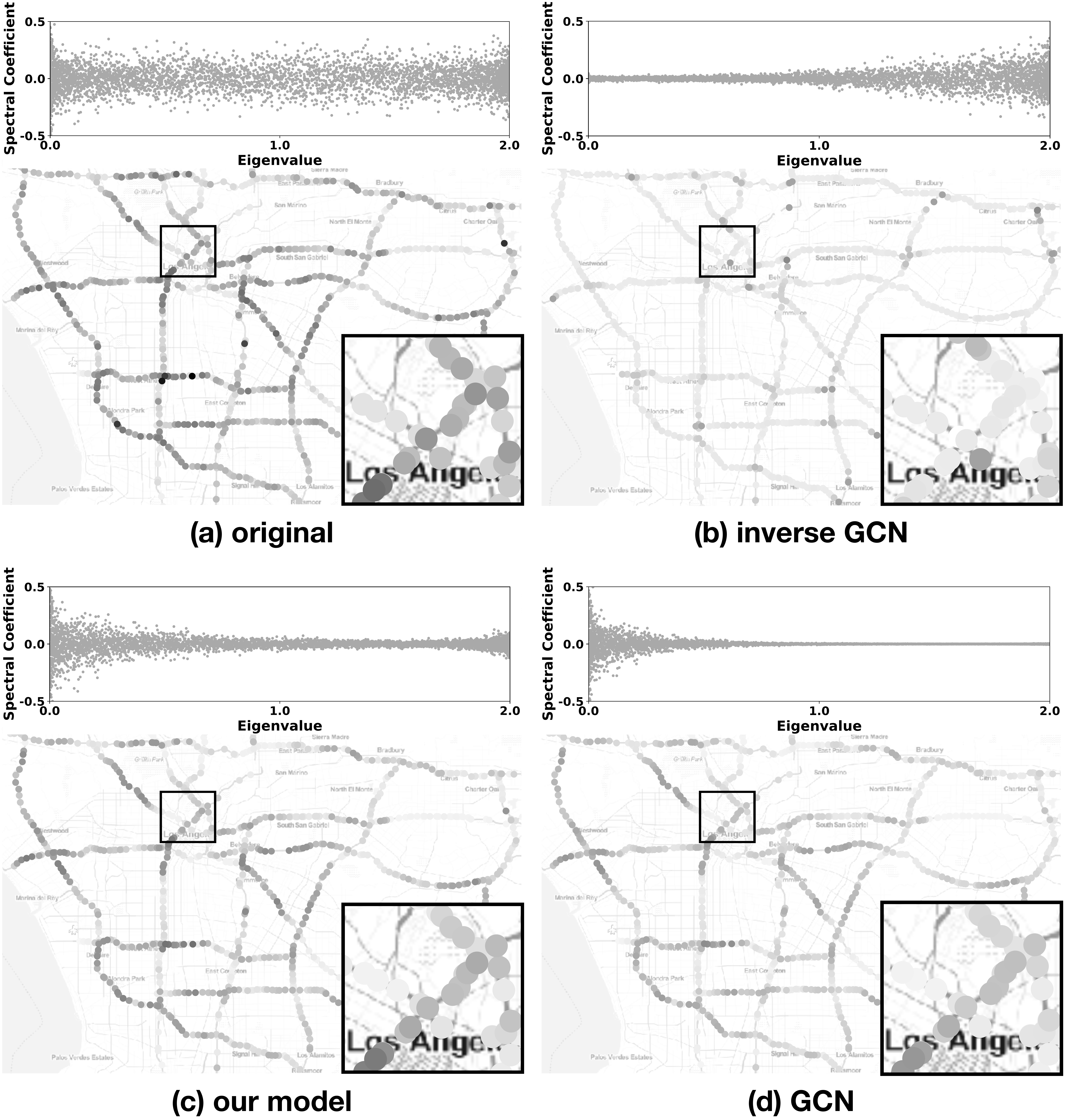}
\end{center}
\caption{Illustration of road occupancy rates decoded by different approaches at 5pm on June 24, 2020 in District 7 of California. A darker color implies a higher road occupancy rate. In each figure above, we plot the reconstructed signal in spatial domain together with coefficients in spectral domain. (a) the original, (b) inverse GCN (RMSE = 0.11), (c) GDN/our model (RMSE = 0.07), (d) GCN (RMSE = 0.09).  }
\label{vis_road}
\end{figure}

To cope with the noise threshold estimation problem, we introduce a parameterized matrix $W_2$ in wavelet domain and eliminate these noise coefficients via a $\text{ReLU}$ function,  in the hope of learning a separating threshold from the data itself. Take the noised representation $M$ raised in Section \ref{inv} into consideration, we derive the reconstructed signal as:
\begin{equation}
 X' = \Psi_s\text{ReLU}(\Psi_s^{-1}MW_4)W_5.
\label{equ.m}
\end{equation}
where $W_4$ and $W_5$ are trainable parameters.

We then contrast the difference between Wavelet Neural Network (WNN) in this work and other related WNNs including GWNN \citep{xu2019graph} and GRAPHWAVE \citep{donnat2018learning}.
\paragraph{Scalability}
An important issue in WNN is that one needs to avoid explicit eigendecomposition to compute wavelet bases for large graphs. 
Both GRAPHWAVE and GWNN, though try to avoid eigendecomposition by exploiting Chebyshev polynomials, still rely integral operations (see Appendix D in \citet{xu2019graph}) and there is no clear way that we can scale up. Differently, we  use Maclaurin series to approximate wavelet bases, which has explicit polynomial coefficients and can resemble the heat kernel well when $n =3$. Please refer to Figure \ref{loworder} for more details.   
\paragraph{De-noising}
The purpose of both GWNN and GRAPHWAVE is to derive node presentations with localized graph convolution and flexible neighborhood, such that downstream tasks like classification can be simplified. On the contrary, our work implements wavelet neural network in the purpose of detaching the useful information and the noise amplified by inverse GCN. Due to the different purpose, our work applies the activation function in wavelet domain while GWNN in the original vertex domain.

\subsection{Visualization}\label{rgd}
In Figure \ref{vis_road}, we illustrate the difference between GDN and each component inside by visualizing the reconstructed road occupancy rates in a traffic network. The traffic network targets District 7 of California collected from Caltrans Performance Measurement
System (PeMS)\footnote{\url{http://pems.dot.ca.gov/}}. We select 4438 sensor stations as the node set $V$ and collect their road average occupancy rates at 5pm on June 24, 2020. Following \citet{li2017diffusion}, we construct an adjacency matrix A by denoting $A_{ij} = 1$ if sensor station $v_j$ and $v_i$ are adjacent on a freeway along the same direction. We reconstruct road occupancy rate of each sensor station with three decoder variants: (b) inverse GCN, (c) GDN, (d) GCN. Here the three variants share the same encoders defined in Section \ref{en}.

As can be seen, while both GDN and GCN decoders can resemble the input signal in low frequency, GDN decoder retains more high frequency information. For inverse GCN decoder, although keeping much high frequency information (mixed with noise), it drops lots of low frequencies, making the decoded signals less similar to the input in Euclidean space.  

\eat{We plot four graphs and their samples generated by DGVAE in Figure \ref{genp} with latent cluster dimension $K=3$. We let the number of edges for graph samples equal with the number for the input graphs. As we have analyzed, most cluster members are connected and uniformly distributed over the three clusters, indicating our model encourages a balanced graph cluster size.
}
\section{Experiments}\label{sec.exp}
We first validate both the proposed graph autoencoder framework and GDN in two tasks including unsupervised graph-level representation \citep{sun2019infograph,icml2020_1971} and social recommendation \citep{jamali2010matrix,monti2017geometric,berg2018graph}, we then test GDN in graph generation tasks \citep{kipf2016variational, grover2018graphite}.

\subsection{unsupervised graph-level representation}
Following recent methods \citep{sun2019infograph,icml2020_1971}, we evaluate the effectiveness of the unsupervised graph-level representations on downstream graph classification tasks.
\paragraph{Data and baselines}
We use five graph classification benchmarks including IMDB-Binary, IMDB-Multi, Reddit-Binary, PROTEINS and DD \citep{NarayananCCLS16,ying2018hierarchical}.  For the detailed statistics, please refer to Table \ref{gcr}.  We compare with six graph kernels: Random Walk (RW) \citep{gartner2003graph}, Shortest Path Kernel (SP) \citep{borgwardt2005shortest}, Graphlet Kernel (GK) \citep{shervashidze2009efficient}, Weisfeiler-Lehman Sub-tree Kernel (WL) \citep{shervashidze2011weisfeiler}, Deep Graph Kernels (DGK) \citep{Yanardag:2015} and Multi-Scale Laplacian Kernel (MLG) \citep{kondor2016multiscale}. In addition, we compare with four unsupervised graph-level representation learning methods: SUB2VEC \citep{adhikari2018sub2vec}, GRAPH2VEC \citep{NarayananCCLS16}, INFOGRAPH \citep{sun2019infograph} and MVGRL \citep{icml2020_1971}. We also include the results of recent supervised graph classification models: GCN \citep{kipf2017semi}, GAT \citep{velivckovic2017graph}, GIN \citep{xu2018powerful}. We denote our framework using (1) GCN \citep{kipf2017semi} in the decoders as ALATION-GCN\footnote{note ALATION-GCN coincides with \citet{zhang2020graph}.}, (2) inverse of GCN in Section \ref{inv} in the decoders as ALATION-INVERSE-GCN. 

\paragraph{Setup}
We adopt the same procedure of previous works \citep{sun2019infograph,icml2020_1971} and report the mean 10-fold cross validation accuracy with standard deviation after 5 runs using LIBSVM \citep{chang2011libsvm}. We report results from previous papers with the same setup if available. For the datasets PROTEINS and DD, we implement strong baselines including INFOGRAPH and MVGRL, with a hyperparameter search according the papers.

We train our model using minibatch based Adam optimizer with a learning rate of 0.01.  We use the cross-entropy loss to reconstruct the features. In our encoders, the best variants of GNN are chosen from GCN \citep{kipf2017semi} and Heatts \citep{li2020heatts}.  For Heatts, we let $s=1$ for all experiments. For detailed hyperparameter settings, please refer to Appendix \ref{p.a}.

\begin{table}[t]
  \caption{Mean 10-fold cross validation accuracy on five graph datasets for kernel and unsupervised methods}
  \begin{center}
  \scalebox{0.77}{
  \begin{tabular}{ccccccc}
    \toprule
    &{Datasets}&\bf IMDB-BIN&\bf IMDB-MULTI&\bf REDDIT-BIN&\bf PROTEINS&\bf DD\\
    \midrule
    \parbox[t]{2mm}{\multirow{6}{*}{\rotatebox[origin=c]{90}{Kernel}}} &RW \citep{gartner2003graph}&50.7 \textpm  0.3&34.7 \textpm 0.2&-&74.2 \textpm 0.4&-\\
	&SP \citep{borgwardt2005shortest}  &55.6 \textpm 0.2& 38.0 \textpm 0.3 &64.1 \textpm 0.1 & 75.07 \textpm 0.5&78.7 \textpm 3.9\\
    &GK \citep{shervashidze2009efficient} &65.9 \textpm 1.0&43.9 \textpm 0.4&77.3 \textpm 0.2&71.67 \textpm 0.6&74.9 \textpm 3.8\\
	&WL \citep{shervashidze2011weisfeiler}&72.3 \textpm 3.4&47.0 \textpm 0.5&68.8 \textpm 0.4&72.92 \textpm 0.6&76.4 \textpm 2.4\\
	&DGK \citep{Yanardag:2015} &67.0 \textpm 0.6&44.6 \textpm 0.5&78.0 \textpm 0.4&75.7 \textpm 0.5&-\\
	&MLG \citep{kondor2016multiscale}&66.6 \textpm 0.3&41.2 \textpm 0.0&-&\bf 76.3 \textpm 0.7&-\\
	\midrule
	\parbox[t]{2mm}{\multirow{4}{*}{\rotatebox[origin=c]{90}{Supervised}}}&GCN \citep{kipf2017semi}&74.0 \textpm 3.4&51.9 \textpm 3.8&50.0 \textpm 0.0&76.0 \textpm 3.2&-\\
	&GAT \citep{velivckovic2017graph}&70.5 \textpm 2.3&47.8 \textpm 3.1&85.2 \textpm 3.3&-&-\\
	&GIN-0 \citep{xu2018powerful}&75.1 \textpm 5.1&\bf 52.3 \textpm 2.8&\textbf{92.4 \textpm 2.5}&76.2 \textpm 2.8&-\\
	&GIN-$\epsilon$ \citep{xu2018powerful}&74.3 \textpm 5.1& 52.1 \textpm 3.6& 92.2 \textpm 2.3&75.9 \textpm 3.8&-\\
	\midrule
	\parbox[t]{2mm}{\multirow{8}{*}{\rotatebox[origin=c]{90}{Unsupervised}}}&VGAE \citep{kipf2016variational} &64.9\textpm 0.38&38.9\textpm 0.46 &- &72.4\textpm 0.42 & 76.3\textpm 0.34 \\
	&SUB2VEC \citep{adhikari2018sub2vec}&55.3 \textpm 1.5&36.7 \textpm 0.8&71.5 \textpm 0.4&-&-\\
	&GRAPH2VEC \citep{NarayananCCLS16}&71.1 \textpm 0.5&50.4 \textpm 0.9&75.8 \textpm 1.0&73.3 \textpm 2.1&-\\
	&INFOGRAPH \citep{sun2019infograph}&73.0 \textpm 0.9&49.7 \textpm 0.5&82.5 \textpm 1.4&75.0 \textpm 1.3&76.9 \textpm 1.3 \\
	&MVGRL \citep{icml2020_1971}&74.2 \textpm 0.7& 51.2 \textpm 0.5& 84.5 \textpm 0.6&75.9 \textpm 1.9&78.3 \textpm 1.7\\
	\cline{2-7}
	&ALATION-INVERSE-GCN&73.2 \textpm 1.9&50.4 \textpm 1.3&83.9 \textpm 1.3&75.0 \textpm 1.2&77.8 \textpm 1.3\\
	&ALATION-GCN&74.2 \textpm 1.4&49.6 \textpm 1.2& 84.6 \textpm 1.2&74.9 \textpm 1.2&77.9 \textpm 1.0\\
	&OURS&\bf 76.0 \textpm 1.3& 51.5 \textpm 1.4& 86.0 \textpm 1.7& 76.1 \textpm 1.4&\bf 79.1 \textpm 1.5\\
  \bottomrule
\end{tabular}
}
\end{center}
 \label{gcr}
\end{table}

\paragraph{Results} The classification accuracies on the five benchmarks are shown in Table \ref{gcr}. MLG, as a kernel method, performs well on PROTEINS. However, it suffers from a long run time and takes more than 1 day on two larger datasets, as observed in INFOGRAPH. Our method achieves the best results in 4 out of 5 datasets compared with both kernel and unsupervised models, e.g., it achieves 1.5\% improvement over previous state-of-the-art on Reddit-Binary, and 1.8\% improvement on IMDB-Binary, which shows the superiority of our methods. When compared with supervised graph classification models, ours beats the best supervised classification model GIN on IMDB-BIN, on-par with GIN on PROTEINS, IMDB-MULTI, and only loses on REDDIT.

\paragraph{Ablation study} We investigate the role of each component in our GDN. We let the three variants share the same depth of layers and parameter size. As observed in Table \ref{gcr}, when decoding by pure inverse of GCN, the performance is just comparable to that of GCN decoders and outperformed by our GDN decoders in all 5 datasets, indicating the effectiveness of this hybrid design.

\paragraph{Running time} We observe that our model runs significantly faster than INFOGRAPH and MVGRL. Our model takes 10s to train one epoch of PORTEINS on Tesla P40 24G, while INFOGRAPH needs 127s and MVGRL needs 193s. This is because our model neglects the tedious process of negative sampling used in both INFOGRAPH and MVGRL.

\subsection{social recommendation}
In our model, we consider social recommendation as feature recovery on graphs, i.e., user-item ratings are represented as the feature matrix and a social network among users is represented as the graph structure. The model is trained on an incomplete version of feature matrix (training data) and is used to infer the potential ratings (test data).

\paragraph{Data and baselines}
We use two social recommendation benchmark datasets: Douban and Ciao\footnote{https://www.ciao.co.uk/}. Both datasets consist of user ratings for items and incorporate a social network among users. For Douban, we use the preprocessed dataset provided by \citet{monti2017geometric}. For Ciao, we use a sub-matrix of 7,317 users and 1,000 items. Dataset statistics are summarized in Table \ref{ssr} in the Appendix. We compare with three baselines: sRGCNN \citep{monti2017geometric}, GC-MC \citep{berg2018graph}, and GraphRec \citep{fan2019graph}. Following \citet{candes2010matrix}, we consider the few available entries (training data) should be corrupted with noise in reality, and add random ratings to the training data at different level $p \in \{0,0.1,\ldots,1\}$, where $p = 0$ denotes no noise is added. 

\begin{table}[t]
  \caption{Comparison of different methods on social recommendation tasks}
  \begin{center}
  \scalebox{0.77}{
  \begin{tabular}{ccccc}
    \toprule
Datasets&\multicolumn{2}{c}{\bf Ciao}&\multicolumn{2}{c}{\bf Douban}\\
\midrule
      -  &ILS&RMSE&ILS&RMSE\\
    \midrule
	sRGCNN \citep{monti2017geometric}&1.63\%&1.183&6.03\%&0.801\\
	GC-MC \citep{berg2018graph} &1.09\%&\bf 1.061&3.90\%&\bf 0.734\\
	GraphRec \citep{fan2019graph} &1.24\%&1.062&8.27\%&0.754\\
	\midrule
	OURS &\bf 0.48\%&1.071&\bf 2.65\%&0.745\\
  \bottomrule
\end{tabular}
}
\end{center}
 \label{gger}
 \vspace{-0.25cm}
\end{table}

\begin{figure}[h]
\begin{center}
\includegraphics [width=0.7\textwidth]{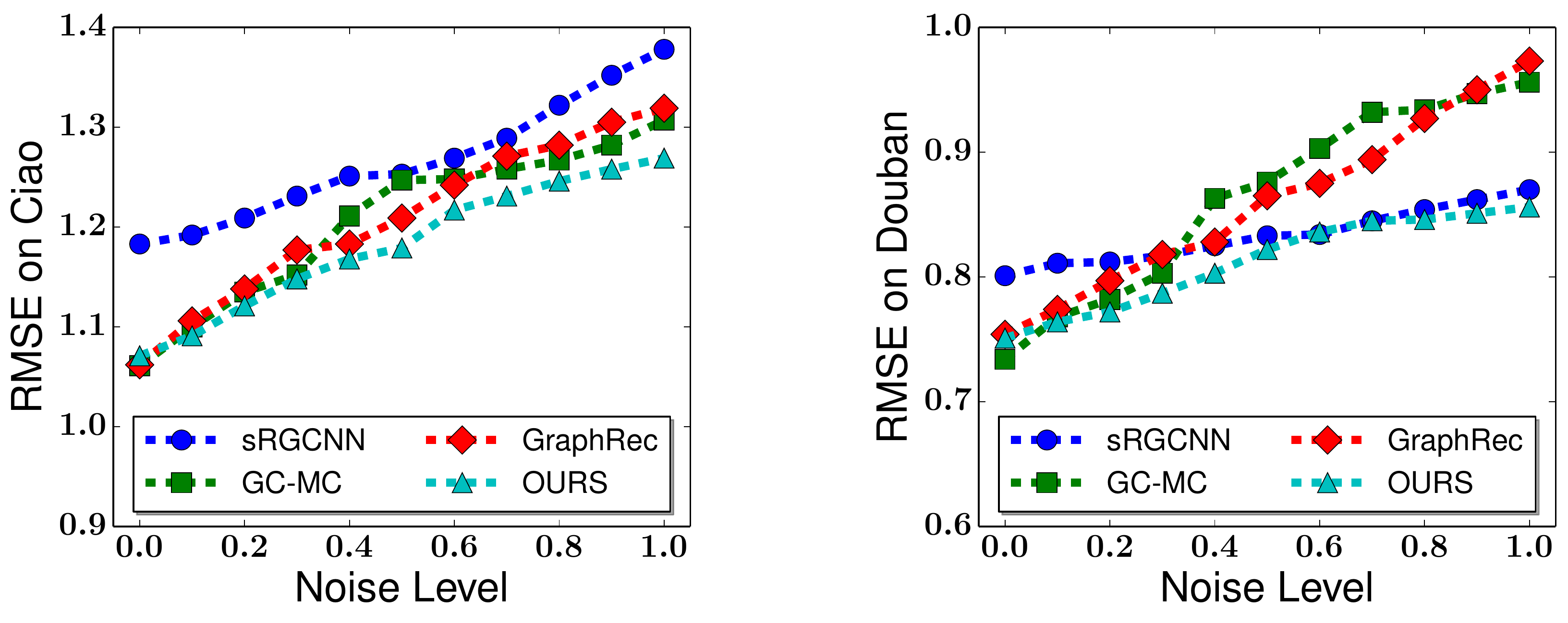}
\end{center}
\caption{Comparison of different methods on social recommendation tasks. The x-axis denotes the noise level while the y-axis denotes RMSE on $\{\text{Ciao}, \text{Douban}\}$.}
\label{noise_level}
\end{figure}

\paragraph{Setup}
For social recommendation, we neglect pooling and unpooling operations as the predictive task is within a \emph{single} graph.  We use the Mean Squared Error (MSE) loss to reconstruct the ratings. Same as in GC-MC \citep{berg2018graph},  we stack the output of the first GCN layer and the second GCN layer in our encoders, and use left normalization to preprocess adjacency matrix. 
For recommendation accuracy, we report Root Mean Squared Error (RMSE). For recommendation diversification \citep{ziegler2005improving}, we report Intra-List Similarity (ILS).  For the definition of ILS and detailed hyperparameter settings, please refer to Appendix \ref{p.a}.

\paragraph{Results} The recommendation RMSE and ILS on the two benchmarks ($p=0$) are shown in Table \ref{gger}. Our method performs on a par with state-of-the-art methods on RMSE, with only lose to the best model GC-MC by 0.01 on Ciao and 0.011 on Douban. As for ILS, our method performs the best, e.g., it beats the second best model GC-MC by 0.61\% on Ciao and 1.25\% on douban, which shows the superiority of our methods on recommendation diversification. The recommendation RMSE with respect to different noise level is shown in Figure \ref{noise_level}. As can be seen, our method performs the best with respect to different noise level ($p > 0$), e.g., it achieves 0.04 improvement over the second best model GC-MC on Ciao when $p = 1$, and 0.014 improvement over the second best model sRGCNN on Douban, which shows the superiority when user-item interactions are noisy. 

\subsection{Graph generation}
\paragraph{Data and baselines}
We use three molecular graphs: MUTAG \citep{kriege2012subgraph} containing mutagenic compounds, PTC-MR \citep{kriege2012subgraph} containing compounds tested for carcinogenicity and ZINC \citep{irwin2012zinc} containing druglike organic molecules, to evaluate the performance of GDN on graph generation.  As our proposed autoencoder framework is not suitable for graph generation task, we test GDN on two popular variational autoencoder framework including VGAE \citep{kipf2016variational} and Graphite \citep{grover2018graphite}. Our purpose is to validate if GDN helps with the generation performance. 
\eat{
\begin{table}[t]
  \caption{The effect of GDN with various graph generation methods}
  \begin{center}
  \scalebox{0.7}{
  \begin{tabular}{ccccccccccc}
    \toprule
Datasets&\multicolumn{3}{c}{\bf MUTAG}&\multicolumn{3}{c}{\bf PTC-MR} &\multicolumn{3}{c}{\bf ZINC}\\
\midrule
      -  &$\log p(A|Z)$&AUC&AP&$\log p(A|Z)$&AUC&AP&$\log p(A|Z)$&AUC&AP\\
    \midrule
	VGAE \citep{kipf2016variational}&-1.156&0.869&0.645&-1.366&0.566&0.433 & -1.033 &0.551 &0.288\\
	VGAE with GDN &-1.114&0.880&0.678& -1.351&0.760&0.602  & -0.997 & \bf 0.862& \bf 0.613\\
	\midrule
	Graphite \citep{grover2018graphite}&-1.140&0.868&0.632&-1.362&0.564&0.437 &-1.043 &0.559&0.288\\
	Graphite with GDN &\bf -1.104&\bf 0.882&\bf 0.681&\bf -1.347&\bf 0.773&\bf 0.613 &\bf -0.989 &0.851&0.593\\
  \bottomrule
\end{tabular}
}
\end{center}
 \label{ggggg}
 \vspace{-0.25cm}
\end{table}
}

\begin{table}[t]
  \caption{The effect of GDN with various graph generation methods}
  \begin{center}
  \scalebox{0.7}{
  \begin{tabular}{ccccccccccc}
    \toprule
Datasets&\multicolumn{3}{c}{\bf MUTAG}&\multicolumn{3}{c}{\bf PTC-MR} &\multicolumn{3}{c}{\bf ZINC}\\
\midrule
      -  &$\log p(A|Z)$&AUC&AP&$\log p(A|Z)$&AUC&AP&$\log p(A|Z)$&AUC&AP\\
    \midrule
	VGAE \citep{kipf2016variational}&-1.156&0.869&0.645&-1.366&0.566&0.433 & -1.035 &0.556 &0.288\\
	VGAE with GDN &-1.114&0.880&0.678& -1.351&0.760&0.602  & -1.006 & \bf 0.858& \bf 0.611\\
	\midrule
	Graphite \citep{grover2018graphite}&-1.140&0.868&0.632&-1.362&0.564&0.437 &-1.039 &0.553&0.288\\
	Graphite with GDN &\bf -1.104&\bf 0.882&\bf 0.681&\bf -1.347&\bf 0.773&\bf 0.613 &\bf -0.998 &0.838&0.567\\
  \bottomrule
\end{tabular}
}
\end{center}
 \label{ggggg}
 \vspace{-0.25cm}
\end{table}

\paragraph{Setup}
For VGAE and Graphite, we reconstruct the graph structures using their default methods and the features using GDN. The two reconstructions share the same encoders and sample from the same latent distributions.  We train for 200 iterations with a learning rate of 0.01.  The output dimension of the first hidden layer is 32 and that of the second-layer is 16. For MUTAG and PTC-MR, we use all the graph samples. We use $50\%$ samples as train set and the remaining $50\%$ as test set. For ZINC, we use the default train-test split (See Table \ref{gcrr} in the Appendix for the details). 
\paragraph{Results} Evaluating the sample quality of generative models is challenging \citep{theis2015note}. In this work, we validate the generative performance with the log-likelihood ($\log p(A|Z)$),  area under the
receiver operating characteristics curve (AUC) and average precision (AP). As shown in Table \ref{ggggg}, GDN can improve the generative performance of both VGAE and Graphite in general, e.g., it improve the AP score of Graphite by $4.9\%$ on MUTAG , the AUC score of VGAE by $19.4\%$ on PTC-MR and $\log p(A|Z)$ of VGAE by $5.4\%$ on ZINC, which shows the superiority of GDN.

\section{Conclusion}\label{con}
In this paper, we present a symmetric graph autoencoder framework in an unsupervised way. The proposed framework relies on Graph Deconvolutional Networks (GDNs), the opposite of GCNs that recover graph signals from smoothed representations. The introduced GDN uses spectral graph convolutions with a \emph{high pass} filter to obtain inversed signals and then de-noises the inversed signals in wavelet domain. The effectiveness of the proposed method is validated on unsupervised graph-level representation, social recommendation and graph generation tasks.

\bibliography{iclr2021_conference}

\begin{thebibliography}{57}
\providecommand{\natexlab}[1]{#1}
\providecommand{\url}[1]{\texttt{#1}}
\expandafter\ifx\csname urlstyle\endcsname\relax
  \providecommand{\doi}[1]{doi: #1}\else
  \providecommand{\doi}{doi: \begingroup \urlstyle{rm}\Url}\fi

\bibitem[Adhikari et~al.(2018)Adhikari, Zhang, Ramakrishnan, and
  Prakash]{adhikari2018sub2vec}
Bijaya Adhikari, Yao Zhang, Naren Ramakrishnan, and B~Aditya Prakash.
\newblock Sub2vec: Feature learning for subgraphs.
\newblock In \emph{Pacific-Asia Conference on Knowledge Discovery and Data
  Mining}, pp.\  170--182. Springer, 2018.

\bibitem[Banham \& Katsaggelos(1997)Banham and Katsaggelos]{banham1997digital}
Mark~R Banham and Aggelos~K Katsaggelos.
\newblock Digital image restoration.
\newblock \emph{IEEE signal processing magazine}, 14\penalty0 (2):\penalty0
  24--41, 1997.

\bibitem[Bengio(2009)]{bengio2009learning}
Yoshua Bengio.
\newblock \emph{Learning deep architectures for AI}.
\newblock Now Publishers Inc, 2009.

\bibitem[Berg et~al.(2018)Berg, Kipf, and Welling]{berg2018graph}
Rianne van~den Berg, Thomas~N Kipf, and Max Welling.
\newblock Graph convolutional matrix completion.
\newblock \emph{SIGKDD}, 2018.

\bibitem[Bianchi et~al.(2020)Bianchi, Grattarola, and
  Alippi]{bianchi2020mincutpool}
Filippo~Maria Bianchi, Daniele Grattarola, and Cesare Alippi.
\newblock Spectral clustering with graph neural networks for graph pooling.
\newblock In \emph{Proceedings of the 37th international conference on Machine
  learning}, pp.\  2729--2738. ACM, 2020.

\bibitem[Borgwardt \& Kriegel(2005)Borgwardt and
  Kriegel]{borgwardt2005shortest}
K.~M. Borgwardt and H.-P. Kriegel.
\newblock Shortest-path kernels on graphs.
\newblock In \emph{ICDM}, pp.\  74--81, 2005.

\bibitem[Candes \& Plan(2010)Candes and Plan]{candes2010matrix}
Emmanuel~J Candes and Yaniv Plan.
\newblock Matrix completion with noise.
\newblock \emph{Proceedings of the IEEE}, 98\penalty0 (6):\penalty0 925--936,
  2010.

\bibitem[Chang \& Lin(2011)Chang and Lin]{chang2011libsvm}
Chih-Chung Chang and Chih-Jen Lin.
\newblock Libsvm: A library for support vector machines.
\newblock \emph{ACM transactions on intelligent systems and technology (TIST)},
  2\penalty0 (3):\penalty0 1--27, 2011.

\bibitem[Chang et~al.(2000)Chang, Yu, and Vetterli]{chang2000adaptive}
S~Grace Chang, Bin Yu, and Martin Vetterli.
\newblock Adaptive wavelet thresholding for image denoising and compression.
\newblock \emph{IEEE transactions on image processing}, 9\penalty0
  (9):\penalty0 1532--1546, 2000.

\bibitem[Defferrard et~al.(2016)Defferrard, Bresson, and
  Vandergheynst]{defferrard2016convolutional}
Micha{\"e}l Defferrard, Xavier Bresson, and Pierre Vandergheynst.
\newblock Convolutional neural networks on graphs with fast localized spectral
  filtering.
\newblock In \emph{Advances in neural information processing systems}, pp.\
  3844--3852, 2016.

\bibitem[Deng et~al.(2020)Deng, Zhao, Wang, Zhang, and Feng]{deng2019graphzoom}
Chenhui Deng, Zhiqiang Zhao, Yongyu Wang, Zhiru Zhang, and Zhuo Feng.
\newblock Graphzoom: A multi-level spectral approach for accurate and scalable
  graph embedding.
\newblock \emph{ICLR}, 2020.

\bibitem[Donnat et~al.(2018)Donnat, Zitnik, Hallac, and
  Leskovec]{donnat2018learning}
Claire Donnat, Marinka Zitnik, David Hallac, and Jure Leskovec.
\newblock Learning structural node embeddings via diffusion wavelets.
\newblock In \emph{ACM SIGKDD Conference on Knowledge Discovery and Data Mining
  (KDD)}, pp.\  1320--1329, 2018.

\bibitem[Donoho \& Johnstone(1994)Donoho and Johnstone]{donoho1994ideal}
David~L Donoho and Jain~M Johnstone.
\newblock Ideal spatial adaptation by wavelet shrinkage.
\newblock \emph{biometrika}, 81\penalty0 (3):\penalty0 425--455, 1994.

\bibitem[Dumoulin \& Visin(2016)Dumoulin and Visin]{dumoulin2016guide}
Vincent Dumoulin and Francesco Visin.
\newblock A guide to convolution arithmetic for deep learning.
\newblock \emph{arXiv preprint arXiv:1603.07285}, 2016.

\bibitem[Fan et~al.(2019)Fan, Ma, Li, He, Zhao, Tang, and Yin]{fan2019graph}
Wenqi Fan, Yao Ma, Qing Li, Yuan He, Eric Zhao, Jiliang Tang, and Dawei Yin.
\newblock Graph neural networks for social recommendation.
\newblock In \emph{The World Wide Web Conference}, pp.\  417--426. ACM, 2019.

\bibitem[Feizi et~al.(2013)Feizi, Marbach, M{\'e}dard, and
  Kellis]{feizi2013network}
Soheil Feizi, Daniel Marbach, Muriel M{\'e}dard, and Manolis Kellis.
\newblock Network deconvolution as a general method to distinguish direct
  dependencies in networks.
\newblock \emph{Nature biotechnology}, 31\penalty0 (8):\penalty0 726--733,
  2013.

\bibitem[Figueiredo \& Nowak(2003)Figueiredo and
  Nowak]{figueiredo2003algorithm}
M{\'a}rio~AT Figueiredo and Robert~D Nowak.
\newblock An em algorithm for wavelet-based image restoration.
\newblock \emph{IEEE Transactions on Image Processing}, 12\penalty0
  (8):\penalty0 906--916, 2003.

\bibitem[Gao \& Ji(2019)Gao and Ji]{gao2019graph}
Hongyang Gao and Shuiwang Ji.
\newblock Graph u-nets.
\newblock In \emph{ICML}, pp.\  2083--2092, 2019.

\bibitem[G{\"a}rtner et~al.(2003)G{\"a}rtner, Flach, and
  Wrobel]{gartner2003graph}
T.~G{\"a}rtner, P.~Flach, and S.~Wrobel.
\newblock On graph kernels: Hardness results and efficient alternatives.
\newblock In \emph{Learning theory and kernel machines}, pp.\  129--143.
  Springer, 2003.

\bibitem[Grover et~al.(2019)Grover, Zweig, and Ermon]{grover2018graphite}
Aditya Grover, Aaron Zweig, and Stefano Ermon.
\newblock Graphite: Iterative generative modeling of graphs.
\newblock In \emph{The International Conference on Machine Learning (ICML)},
  pp.\  2434--2444, 2019.

\bibitem[Hammond et~al.(2011)Hammond, Vandergheynst, and
  Gribonval]{hammond2011wavelets}
David~K Hammond, Pierre Vandergheynst, and R{\'e}mi Gribonval.
\newblock Wavelets on graphs via spectral graph theory.
\newblock \emph{Applied and Computational Harmonic Analysis}, 30\penalty0
  (2):\penalty0 129--150, 2011.

\bibitem[Hassani \& Khasahmadi(2020)Hassani and Khasahmadi]{icml2020_1971}
Kaveh Hassani and Amir~Hosein Khasahmadi.
\newblock Contrastive multi-view representation learning on graphs.
\newblock In \emph{Proceedings of International Conference on Machine
  Learning}, pp.\  3451--3461. 2020.

\bibitem[Irwin et~al.(2012)Irwin, Sterling, Mysinger, Bolstad, and
  Coleman]{irwin2012zinc}
John~J Irwin, Teague Sterling, Michael~M Mysinger, Erin~S Bolstad, and Ryan~G
  Coleman.
\newblock Zinc: a free tool to discover chemistry for biology.
\newblock \emph{Journal of chemical information and modeling}, 52\penalty0
  (7):\penalty0 1757--1768, 2012.

\bibitem[Jamali \& Ester(2010)Jamali and Ester]{jamali2010matrix}
Mohsen Jamali and Martin Ester.
\newblock A matrix factorization technique with trust propagation for
  recommendation in social networks.
\newblock In \emph{RecSys}, pp.\  135--142, 2010.

\bibitem[Kipf \& Welling(2016)Kipf and Welling]{kipf2016variational}
Thomas~N Kipf and Max Welling.
\newblock Variational graph auto-encoders.
\newblock In \emph{Conference on Neural Information Processing Systems
  (NeurIPS) Workshop on Bayesian Deep Learning}, 2016.

\bibitem[Kipf \& Welling(2017)Kipf and Welling]{kipf2017semi}
Thomas~N. Kipf and Max Welling.
\newblock Semi-supervised classification with graph convolutional networks.
\newblock In \emph{The International Conference on Learning Representations
  (ICLR)}, 2017.

\bibitem[Kondor \& Pan(2016)Kondor and Pan]{kondor2016multiscale}
Risi Kondor and Horace Pan.
\newblock The multiscale laplacian graph kernel.
\newblock In \emph{Advances in Neural Information Processing Systems}, pp.\
  2990--2998, 2016.

\bibitem[Kriege \& Mutzel(2012)Kriege and Mutzel]{kriege2012subgraph}
Nils Kriege and Petra Mutzel.
\newblock Subgraph matching kernels for attributed graphs.
\newblock \emph{ICML}, pp.\  291--298, 2012.

\bibitem[Kundur \& Hatzinakos(1996)Kundur and Hatzinakos]{kundur1996blind}
Deepa Kundur and Dimitrios Hatzinakos.
\newblock Blind image deconvolution.
\newblock \emph{IEEE signal processing magazine}, 13\penalty0 (3):\penalty0
  43--64, 1996.

\bibitem[Lee et~al.(2019)Lee, Lee, and Kang]{lee2019self}
Junhyun Lee, Inyeop Lee, and Jaewoo Kang.
\newblock Self-attention graph pooling.
\newblock \emph{Proceedings of the 36th International Conference on Machine
  Learning (ICML)}, 2019.

\bibitem[Li et~al.(2019)Li, Rong, Cheng, Meng, Huang, and Huang]{jiawww19}
Jia Li, Yu~Rong, Hong Cheng, Helen Meng, Wenbing Huang, and Junzhou Huang.
\newblock Semi-supervised graph classification: A hierarchical graph
  perspective.
\newblock In \emph{The World Wide Web Conference (WWW)}, pp.\  972–982, 2019.

\bibitem[Li et~al.(2020)Li, Yu, Li, Zhang, Zhao, Rong, Cheng, and
  Huang]{li2020heatts}
Jia Li, Jianwei Yu, Jiajin Li, Honglei Zhang, Kangfei Zhao, Yu~Rong, Hong
  Cheng, and Junzhou Huang.
\newblock Dirichlet graph variational autoencoder.
\newblock In \emph{Neurips}, 2020.

\bibitem[Li et~al.(2018)Li, Yu, Shahabi, and Liu]{li2017diffusion}
Yaguang Li, Rose Yu, Cyrus Shahabi, and Yan Liu.
\newblock Diffusion convolutional recurrent neural network: Data-driven traffic
  forecasting.
\newblock \emph{ICLR}, 2018.

\bibitem[Long et~al.(2015)Long, Shelhamer, and Darrell]{long2015fully}
Jonathan Long, Evan Shelhamer, and Trevor Darrell.
\newblock Fully convolutional networks for semantic segmentation.
\newblock In \emph{Proceedings of the IEEE conference on computer vision and
  pattern recognition}, pp.\  3431--3440, 2015.

\bibitem[Monti et~al.(2017)Monti, Bronstein, and Bresson]{monti2017geometric}
Federico Monti, Michael Bronstein, and Xavier Bresson.
\newblock Geometric matrix completion with recurrent multi-graph neural
  networks.
\newblock In \emph{Neurips}, pp.\  3697--3707, 2017.

\bibitem[Narayanan et~al.(2017)Narayanan, Chandramohan, Chen, Liu, and
  Saminathan]{NarayananCCLS16}
Annamalai Narayanan, Mahinthan Chandramohan, Lihui Chen, Yang Liu, and
  Santhoshkumar Saminathan.
\newblock subgraph2vec: Learning distributed representations of rooted
  sub-graphs from large graphs.
\newblock \emph{CoRR}, abs/1606.08928, 2017.

\bibitem[Neelamani et~al.(2004)Neelamani, Choi, and
  Baraniuk]{neelamani2004forward}
Ramesh Neelamani, Hyeokho Choi, and Richard Baraniuk.
\newblock Forward: Fourier-wavelet regularized deconvolution for
  ill-conditioned systems.
\newblock \emph{IEEE Transactions on signal processing}, 52\penalty0
  (2):\penalty0 418--433, 2004.

\bibitem[Noh et~al.(2015)Noh, Hong, and Han]{noh2015learning}
Hyeonwoo Noh, Seunghoon Hong, and Bohyung Han.
\newblock Learning deconvolution network for semantic segmentation.
\newblock In \emph{Proceedings of the IEEE international conference on computer
  vision}, pp.\  1520--1528, 2015.

\bibitem[Shervashidze et~al.(2009)Shervashidze, Vishwanathan, Petri, Mehlhorn,
  and Borgwardt]{shervashidze2009efficient}
N.~Shervashidze, S.V.N. Vishwanathan, T.~Petri, K.~Mehlhorn, and K.~M.
  Borgwardt.
\newblock Efficient graphlet kernels for large graph comparison.
\newblock In \emph{AISTATS}, pp.\  488--495, 2009.

\bibitem[Shervashidze et~al.(2011)Shervashidze, Schweitzer, v.~Leeuwen,
  Mehlhorn, and Borgwardt]{shervashidze2011weisfeiler}
N.~Shervashidze, P.~Schweitzer, E.~J. v.~Leeuwen, K.~Mehlhorn, and K.~M.
  Borgwardt.
\newblock Weisfeiler-lehman graph kernels.
\newblock \emph{Journal of Machine Learning Research}, 12\penalty0
  (Sep):\penalty0 2539--2561, 2011.

\bibitem[Shuman et~al.(2013)Shuman, Narang, Frossard, Ortega, and
  Vandergheynst]{shuman2013emerging}
David~I Shuman, Sunil~K Narang, Pascal Frossard, Antonio Ortega, and Pierre
  Vandergheynst.
\newblock The emerging field of signal processing on graphs: Extending
  high-dimensional data analysis to networks and other irregular domains.
\newblock \emph{IEEE signal processing magazine}, 30\penalty0 (3):\penalty0
  83--98, 2013.

\bibitem[Simonovsky \& Komodakis(2018)Simonovsky and
  Komodakis]{simonovsky2018graphvae}
Martin Simonovsky and Nikos Komodakis.
\newblock Graphvae: Towards generation of small graphs using variational
  autoencoders.
\newblock In \emph{International Conference on Artificial Neural Networks},
  pp.\  412--422. Springer, 2018.

\bibitem[Sun et~al.(2020)Sun, Hoffmann, Verma, and Tang]{sun2019infograph}
Fan-Yun Sun, Jordan Hoffmann, Vikas Verma, and Jian Tang.
\newblock Infograph: Unsupervised and semi-supervised graph-level
  representation learning via mutual information maximization.
\newblock \emph{ICLR}, 2020.

\bibitem[Theis et~al.(2016)Theis, Oord, and Bethge]{theis2015note}
Lucas Theis, A{\"a}ron van~den Oord, and Matthias Bethge.
\newblock A note on the evaluation of generative models.
\newblock \emph{ICLR}, 2016.

\bibitem[Veli{\v{c}}kovi{\'c} et~al.(2018)Veli{\v{c}}kovi{\'c}, Cucurull,
  Casanova, Romero, Lio, and Bengio]{velivckovic2017graph}
Petar Veli{\v{c}}kovi{\'c}, Guillem Cucurull, Arantxa Casanova, Adriana Romero,
  Pietro Lio, and Yoshua Bengio.
\newblock Graph attention networks.
\newblock \emph{ICLR}, 2018.

\bibitem[Veli{\v{c}}kovi{\'{c}} et~al.(2019)Veli{\v{c}}kovi{\'{c}}, Fedus,
  Hamilton, Li{\`{o}}, Bengio, and Hjelm]{velickovic2018deep}
Petar Veli{\v{c}}kovi{\'{c}}, William Fedus, William~L. Hamilton, Pietro
  Li{\`{o}}, Yoshua Bengio, and R~Devon Hjelm.
\newblock {Deep Graph Infomax}.
\newblock In \emph{International Conference on Learning Representations}, 2019.

\bibitem[Vincent et~al.(2010)Vincent, Larochelle, Lajoie, Bengio, Manzagol, and
  Bottou]{vincent2010stacked}
Pascal Vincent, Hugo Larochelle, Isabelle Lajoie, Yoshua Bengio, Pierre-Antoine
  Manzagol, and L{\'e}on Bottou.
\newblock Stacked denoising autoencoders: Learning useful representations in a
  deep network with a local denoising criterion.
\newblock \emph{Journal of machine learning research}, 11\penalty0 (12), 2010.

\bibitem[Wu et~al.(2019)Wu, Souza, Zhang, Fifty, Yu, and
  Weinberger]{pmlr-v97-wu19e}
Felix Wu, Amauri Souza, Tianyi Zhang, Christopher Fifty, Tao Yu, and Kilian
  Weinberger.
\newblock Simplifying graph convolutional networks.
\newblock In \emph{Proceedings of the 36th International Conference on Machine
  Learning (ICML)}, pp.\  6861--6871. PMLR, 2019.

\bibitem[Xu et~al.(2019{\natexlab{a}})Xu, Shen, Cao, Qiu, and
  Cheng]{xu2019graph}
Bingbing Xu, Huawei Shen, Qi~Cao, Yunqi Qiu, and Xueqi Cheng.
\newblock Graph wavelet neural network.
\newblock \emph{The International Conference on Learning Representations
  (ICLR)}, 2019{\natexlab{a}}.

\bibitem[Xu et~al.(2019{\natexlab{b}})Xu, Hu, Leskovec, and
  Jegelka]{xu2018powerful}
Keyulu Xu, Weihua Hu, Jure Leskovec, and Stefanie Jegelka.
\newblock How powerful are graph neural networks?
\newblock \emph{ICLR}, 2019{\natexlab{b}}.

\bibitem[Yanardag \& Vishwanathan(2015)Yanardag and
  Vishwanathan]{Yanardag:2015}
P.~Yanardag and S.V.N. Vishwanathan.
\newblock Deep graph kernels.
\newblock In \emph{KDD}, pp.\  1365--1374, 2015.

\bibitem[Yang \& Segarra(2018)Yang and Segarra]{yang2018enhancing}
Jingkang Yang and Santiago Segarra.
\newblock Enhancing geometric deep learning via graph filter deconvolution.
\newblock In \emph{2018 IEEE Global Conference on Signal and Information
  Processing (GlobalSIP)}, pp.\  758--762. IEEE, 2018.

\bibitem[Ying et~al.(2018)Ying, You, Morris, Ren, Hamilton, and
  Leskovec]{ying2018hierarchical}
Zhitao Ying, Jiaxuan You, Christopher Morris, Xiang Ren, Will Hamilton, and
  Jure Leskovec.
\newblock Hierarchical graph representation learning with differentiable
  pooling.
\newblock In \emph{Advances in neural information processing systems}, pp.\
  4800--4810, 2018.

\bibitem[Zeiler \& Fergus(2014)Zeiler and Fergus]{zeiler2014visualizing}
Matthew~D Zeiler and Rob Fergus.
\newblock Visualizing and understanding convolutional networks.
\newblock In \emph{European conference on computer vision}, pp.\  818--833.
  Springer, 2014.

\bibitem[Zeiler et~al.(2010)Zeiler, Krishnan, Taylor, and
  Fergus]{zeiler2010deconvolutional}
Matthew~D Zeiler, Dilip Krishnan, Graham~W Taylor, and Rob Fergus.
\newblock Deconvolutional networks.
\newblock In \emph{2010 IEEE Computer Society Conference on computer vision and
  pattern recognition}, pp.\  2528--2535. IEEE, 2010.

\bibitem[Zhang et~al.(2020)Zhang, Hu, Yang, Chen, and Yao]{zhang2020graph}
Chun-Yang Zhang, Junfeng Hu, Lin Yang, CL~Philip Chen, and Zhiliang Yao.
\newblock Graph deconvolutional networks.
\newblock \emph{Information Sciences}, 518:\penalty0 330--340, 2020.

\bibitem[Ziegler et~al.(2005)Ziegler, McNee, Konstan, and
  Lausen]{ziegler2005improving}
Cai-Nicolas Ziegler, Sean~M McNee, Joseph~A Konstan, and Georg Lausen.
\newblock Improving recommendation lists through topic diversification.
\newblock In \emph{Proceedings of the 14th international conference on World
  Wide Web}, pp.\  22--32, 2005.

\end{thebibliography}
\bibliographystyle{iclr2021_conference}
\clearpage
\appendix

\begin{table}[t]
  \caption{Statistics of the datasets used in graph classification}
  \begin{center}
  \scalebox{0.77}{
  \begin{tabular}{cccccccc}
    \toprule
    {Datasets}&\bf IMDB-BIN&\bf IMDB-MULTI&\bf REDDIT-BIN&\bf PROTEINS&\bf DD\\
    \midrule
	{(No.Graphs)} &1000& 1500 &2000 & 1113 &1178 \\
	{(No.Classes)}&2&3&2&2&2\\
	{(Avg.Nodes)}&19.8&13.0&508.5&39.1&284.3\\
	{(Avg.Edges)}&193.1&65.9&497.8&72.8&715.7\\
  \bottomrule
\end{tabular}
}
\end{center}
 \label{gcr}
\end{table}

\begin{table}[t]
  \caption{Statistics of the datasets used in social recommendation}
  \begin{center}
  \scalebox{0.85}{
  \begin{tabular}{ccccccc}
    \toprule
Dataset&\bf Users&\bf Items&\bf Train/Test Ratings&\bf Rating Density&\bf Social Connections&\bf Social Density\\
	\midrule
	Douban&3,000&3,000&123,202/13,689&1.52\%&2,690&0.03\%\\
	Ciao&7,317 &1,000 &39,279/16,892&0.77\%&111,781&0.21\%\\
  \bottomrule
\end{tabular}
}
\end{center}
 \label{ssr}
\end{table}

\begin{table}[t]
  \caption{Statistics of the datasets used in graph generation}
  \begin{center}
  \scalebox{0.77}{
  \begin{tabular}{cccc}
    \toprule
    {Datasets}&\bf MUTAG&\bf PTC-MR &\bf ZINC\\
    \midrule
	{(No.Graphs)} &188& 344&250k\\
	{(Avg.Nodes)}&17.9&14.3&23.1\\
	{(Avg.Edges)}&19.8&14.7&23.9\\
  \bottomrule
\end{tabular}
}
\end{center}
 \label{gcrr}
\end{table}
\section{Detailed Model Configuration}\label{p.a}
For the unsupervised graph-level representation experiments, we use the original graph structure in the decoders of the proposed graph autoencoder framework.  The C parameter of LIBSVM \citep{chang2011libsvm} is selected from $\{10^{-3},10^{-2},\ldots,10^2,10^3\}$. The depth of GNN layers in the encoders is set to 2. The output dimension of the first layer is chosen from $\{64,128,256\}$. The output dimension of the second layer is chosen from $\{16,32\}$. The number of clusters $K$ is chosen from $\{16,32\}$. The number of epochs is chosen from $[15,20]$. The batch size is set to 1. We set $\lambda_A = 0$ and $\lambda_X = 1$. We don't use Dropout as it does not improve the performance. For all methods, the embedding dimension is set to 512 and parameters of downstream classifiers are independently tuned using cross validation on training folds of data, in order to have a fair comparison with previous works. The best average classification accuracy is reported for all methods.

For social recommendation, we train our model using Adam optimizer with a learning rate of $\{0.005,0.002\}$. We use the original graph structure in the decoders of the proposed graph autoencoder framework. The output dimension of the first layer is 256. The output dimension of the second layer is 128. The number of epochs is 200. We use full-batch size. We don't use Dropout as it does not improve the performance. Intra-List Similarity (ILS) of one user is defined as:
\begin{equation}
 \text{ILS}_u = \frac{1}{2}\sum_{i_m \in L}\sum_{i_n \in L}\text{sim}(i_m,i_n),
\end{equation}
where $L$ is the item list for the user and it is equal with the top-10 items in this work. $\text{sim}(\cdot, \cdot)$ is the similarity function and is set to cosine similarity. For the representation of items, we use the input user rating vectors.

\section{More related works}\label{p.b}

\paragraph{Unsupervised graph-level representations}
Unsupervised graph-level representation techniques can be generally classified into three groups. The first group is graph kernels. Representative works include Random Walk \citep{gartner2003graph}, Shortest Path Kernel \citep{borgwardt2005shortest}, Graphlet Kernel \citep{shervashidze2009efficient}, Weisfeiler-Lehman Sub-tree Kernel \citep{shervashidze2011weisfeiler}, Deep Graph Kernels \citep{Yanardag:2015} and Multi-Scale Laplacian Kernel \citep{kondor2016multiscale}.  This group of techniques works on finding informative sub-structures and computes similarities based on these sub-structures. The second group is contrastive methods.  Representative works include SUB2VEC \citep{adhikari2018sub2vec}, GRAPH2VEC \citep{NarayananCCLS16}, INFOGRAPH \citep{sun2019infograph} and MVGRL \citep{icml2020_1971}. This group employs a scoring function, e.g., mutual information \citep{velickovic2018deep}, to train a model that can increase the score on positive examples and decrease the score on negative samples. It thus depends on the sophisticated positive/negative sampling strategy to boost the performance. The third group is graph autoencoders \citep{kipf2016variational,grover2018graphite}. Due to the lack of powerful decoders, the performance of this kind is still underestimated.

\paragraph{Social recommendation}
Social recommendations assume a social network among users and these user-user interactions are helpful in boosting
the recommendation performance \citep{jamali2010matrix}. As GNNs have been proven to be capable of learning on graph data, recently recommendation methods based on GNNs have shown impressive results on recommendation accuracy. Representative works include sRGCNN \citep{monti2017geometric}, GC-MC \citep{berg2018graph}, and GraphRec \citep{fan2019graph}. Specifically, GC-MC \citep{berg2018graph} models recommendations as link predictions using graph autoencoder. Though successful, GC-MC uses an MLP decoder and fails to deal with noisy ratings \citep{candes2010matrix}.

\section{Derivative of Inverse GCN}\label{p.a}
\begin{align}
    g_{c}^{-1}(\lambda_i) &= \frac{1}{1-\lambda_{i}} = \sum_{n=0}^{\infty} \frac{\big{(}\frac{1}{1-\lambda_{i}}\big{)}_{\lambda_i=0}^{(n)}}{n!}\lambda_{i}^{n} \\
    & = \sum_{n=0}^{\infty}\frac{(-1)^{n}n!(-1)^n}{n!}\lambda_{i}^{n}=\sum_{n=0}^{\infty} \lambda_{i}^{n} \nonumber
\end{align}
where $(n)$ denotes the $n$-th order derivative. Thus, Eq.(8) can be obtained as:
\begin{align}
    g_{c}^{-1} * x = U\text{diag}(\sum_{n=0}^{\infty} \lambda_1^n,\ldots,\sum_{n=0}^{\infty}\lambda_N^n)U^\top x=U \sum_{n=0}^{\infty} \Lambda^n U^\top x
\end{align}

\end{document}